\def\year{2026}\relax
\documentclass[letterpaper]{article}
\usepackage{microtype}
\usepackage{aaai2026}
\usepackage{times}
\usepackage{helvet}
\usepackage{courier}
\usepackage{float} 
\usepackage{graphicx}
\usepackage{subcaption}
\usepackage{booktabs}
\usepackage{siunitx}
\usepackage{amsmath,amssymb}
\usepackage{natbib}
\usepackage{url}
\usepackage{xspace}
\usepackage{microtype}
\usepackage{makecell}
\usepackage{placeins}
\usepackage[font=small,labelfont=bf]{caption}

\graphicspath{{figs/}{figures/}{images/}}
\title{}
\author{}
\nocite{Sharma2025SDSA}

\usepackage{enumitem}
\setlist{leftmargin=*, itemsep=0.25em, topsep=0.25em}
\title{Towards Fine-Tuning-Based Site Calibration for Knowledge-Guided Machine Learning: A Summary of Results}
\author {
    Ruolei Zeng\textsuperscript{\rm 1},
    Arun Sharma\textsuperscript{\rm 1},
    Shuai An\textsuperscript{\rm 1},
    Mingzhou Yang\textsuperscript{\rm 1},
    Shengya Zhang\textsuperscript{\rm 1},
    Licheng Liu\textsuperscript{\rm 2},
    David Mulla\textsuperscript{\rm 2},
    Shashi Shekhar\textsuperscript{\rm 1}
}
\affiliations {
    \textsuperscript{\rm 1}Department of Computer Science, University of Minnesota, Twin Cities, USA\\
    \textsuperscript{\rm 2}Department of Soil, Water, and Climate, University of Minnesota, Twin Cities, USA\\

    \{zeng0208, arunshar, an000033, yang7492, zhan9051, lichengl, mulla003, shekhar\}@umn.edu
}

\begin{document}

\maketitle
\begin{abstract}
Accurate and cost-effective quantification of the agroecosystem carbon cycle at decision-relevant scales is essential for climate mitigation and sustainable agriculture.  However, both transfer learning and spatial variability exploitation in this field are challenging, as they involve heterogeneous data and complex cross-scale dependencies. Conventional approaches often rely on location-independent parameterizations and independent training, underutilizing transfer learning and spatial heterogeneity in the inputs, and limiting their applicability in regions with strong variability. We propose FTBSC-KGML (Fine-Tuning-Based Site Calibration–Knowledge-Guided Machine Learning), a pretraining- and fine-tuning-based, spatial-variability-aware, and knowledge-guided machine learning framework that augments KGML-ag with a pretraining-fine-tuning process and site-specific parameters. Using a pretraining-fine-tuning process with remote-sensing GPP, climate, and soil covariates collected across multiple midwestern sites, FTBSC-KGML estimates land emissions while leveraging transfer learning and spatial heterogeneity. A key component is a spatial heterogeneity-aware transfer-learning scheme: a globally pretrained model that is fine-tuned per state/site to learn place-aware representations, improving local accuracy under limited data without sacrificing interpretability. Empirically, FTBSC-KGML achieves lower validation error and more consistent explanatory power than a purely global model, better capturing spatial variability across states. This work extends the prior SDSA-KGML framework.
\end{abstract}





%
%





\section{Introduction}

Purely data-driven machine learning (ML) models often achieve limited success in scientific domains due to their high data requirements and inability to produce physically consistent results \cite{10.1145/3514228}. Thus, research communities have begun to explore integrating scientific knowledge with ML in a synergistic manner. The burgeoning field of knowledge-guided machine learning (KGML) offers a promising framework that integrates the strengths of process-based (PB) models, machine learning, and multi-source datasets. KGML has proven effective in spatial prediction tasks, such as land emissions estimation. However, current KGML models use location-independent parameters that overlook spatial heterogeneity across their large footprints, leaving little room for effective global-to-local transfer. As a result, performance and interpretability degrade in settings where processes are spatially variable \cite {Moran1950_SpatialAutocorrelation}. To address these issues, this paper proposes a Fine-Tuning-Based Site Calibration–Knowledge-Guided Machine Learning (FTBSC-KGML) framework as a general schema to enhance current KGML methods by incorporating location-based parameter values 
and cross-site transfer learning. Building on prior work such as SDSA-KGML \cite{Sharma2025SDSA},  the proposed approach introduces a transfer-learning mechanism in which a globally pretrained, physics-guided model is fine-tuned for each site or state. This design bridges awareness of spatial variability with model calibration efficiency, leveraging knowledge from aggregated multi-state training while retaining site-specific interpretability. 

\begin{figure}[t]
\centering
\begin{tabular}{cc}
\hspace*{-1.0em}
\subfloat[Model Input]{
    \includegraphics[width=0.55\linewidth, height=0.45\linewidth]{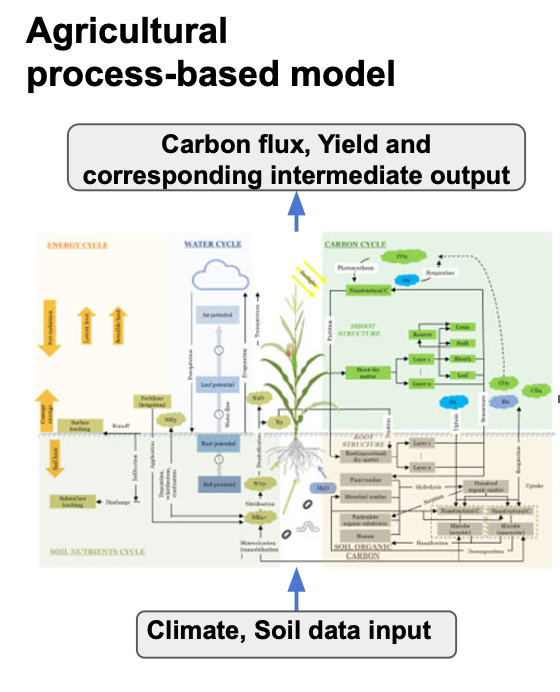}
    \label{fig:input}
}
&
\hspace*{-1.0em}
\subfloat[Model Output]{
    \includegraphics[width=0.55\linewidth, height=0.25\linewidth]{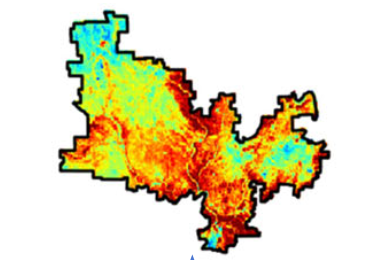}
    \label{fig:output}
}
\end{tabular}

\vspace{-0.4em}
\caption{Input and output of KGML-ag~\cite{Liu2024_KGML_Carbon}.}
\label{fig:overall}
\end{figure}

The problem is important for accurately predicting carbon and other emissions from land-use activities (e.g., agriculture, deforestation). Quantifying and controlling these emissions is crucial for climate change mitigation, optimum crop management, and maintaining sustainable agriculture.

Predicting land emissions is challenging due to the heterogeneity of factors that affect them. This spatial variability, as implied by Tobler’s First Law of Geography \cite{Tobler1970}, encompasses variations in soil characteristics, moisture content, and other environmental conditions. Moreover, collecting ground truth data for this task is costly, complicating the training of large deep learning models. These challenges call for methods that both effectively capture spatial variability and are guided by physical knowledge.

\textbf{Related Work:} The most common approach for predicting land emissions is based on process-based models, which use
scientific theories that accurately explain the phenomena that occur, obeying principles such as mass and energy conservation. However, these models do not perform well in high spatial heterogeneity and variance, which is common in real-world settings \cite {gupta2021spatial}. Other approaches considered, especially for small-area estimation, include data-driven machine learning models. However, these models usually require extensive training data, which can be time-consuming and sometimes impossible to achieve. 

Knowledge-Guided Machine Learning (KGML) methods have been further explored, incorporating elements of both process-based models and data-driven machine learning approaches. For instance, KGML-ag \cite{Liu2024_KGML_Carbon} integrates several pretraining steps with knowledge from \textit{ecosys}, a process-based model for agroecosystems, into a deep learning architecture. KGML-ag effectively addresses challenges such as spatial autocorrelation and scalability to larger datasets. However, as mentioned earlier, the lack of awareness of spatial variability in KGML-ag limits its performance and interpretability.

To address this limitation, \cite{Sharma2025SDSA} proposed the \textit{Spatial Distribution-Shift Aware Knowledge-Guided Machine Learning} (SDSA-KGML) framework, which introduced location-dependent parameters to explicitly account for spatial heterogeneity and distribution shifts across regions such as Illinois, Iowa, and Indiana. Their results demonstrated that incorporating region-specific parameters enhanced local accuracy under strong spatial variability. Building upon this direction, our proposed \textbf{Fine-Tuning-Based Site Calibration} (FTBSC-KGML) extends SDSA-KGML by introducing a transfer-learning mechanism in which a globally pretrained model is fine-tuned for each site or state. This approach bridges spatial variability awareness with model calibration efficiency, leveraging knowledge from aggregated multi-state training while
retaining site-specific interpretability. Empirically, this mechanism significantly improves local performance in data-limited regions, mitigating overfitting and preserving knowledge-guided physical constraints.

Organization: The paper is organized as follows: Section 2 introduces basic concepts. Section 3 formally defines the problem. Section 4 discusses design decisions. Section 5 presents the proposed approach. Section 6 discusses experimental evaluation. Section 7 concludes the paper and discusses future work.

\section{Basic Concepts}
\textbf{Knowledge-Guided Machine Learning-ag:} The KGML-ag model takes climate and soil data as inputs and generates predictions of agricultural carbon fluxes, crop yields, and changes in soil carbon stocks (i.e., carbon quantity) as outputs \cite{fang2018modeling}. The input data are gathered from diverse sources, including Eddy Covariance (EC) flux tower sites \cite{Pastorello2020_FLUXNET2015}, regional survey yield data, remotely sensed gross primary production data, and synthetic data generated by a process-based model. While these data offer a rich set of features, they also present challenges, such as inconsistencies between real and synthetic data and varying data quality \cite{CressieWikle2011_SpatioTemporal}. KGML-ag uses a hierarchical structure with five submodules, including (1) a GRU\_Ra module for daily Autotrophic Respiration (Ra) estimation, (2) a GRU\_Rh module for daily Heterotrophic Respiration (Rh)  estimation, (3) a GRU\_NEE module for daily Net Ecosystem Exchange (NEE) estimation, (4) an attention module for crop yield estimation, and (5) a GRU\_Basis module to connect and support the other four modules \cite{Liu2024_KGML_Carbon}.

\textbf{Synthetic Data:} Along with real sample data, the KGML-ag model is trained on synthetic data generated by the \textit{ecosys} model \cite{Grant2001Ecosys} to improve its generalization. The \textit{ecosys} model simultaneously simulates carbon, water, and nutrient cycles within the soil and plant system based on biophysical and biochemical principles \cite{shang2017spatiotemporal}. In this context, the model is used to generate county-level synthetic data on ecosystem carbon allocation, associated fluxes, and environmental responses. 

\textbf{The 5 Step Training Process:} KGML-ag uses a five-step training method: 1) Pretrain the yield and Ra submodules using the synthetic data. 2) Pretrains the Ra, Rh, and NEE submodules using the synthetic data. After steps 1 and 2, the KGML-ag model can imitate the \textit{ecosys} model for yield, Ra, Rh, and NEE. 3) The yield module is fine-tuned with county-level crop yield data. 4) Use synthetic data to maintain Ra, Rh, and NEE pretraining after yield fine-tuning. Despite the similarity between 2) and 4), experimental results show that they are essential for maintaining predictive accuracy. 5) Ra, Rh, and NEE are fine-tuned using observed data from Eddy Covariance (EC) flux tower sites across the U.S. Midwest. 

\begin{figure}[t]
    \centering
    \includegraphics[width=0.95\linewidth]{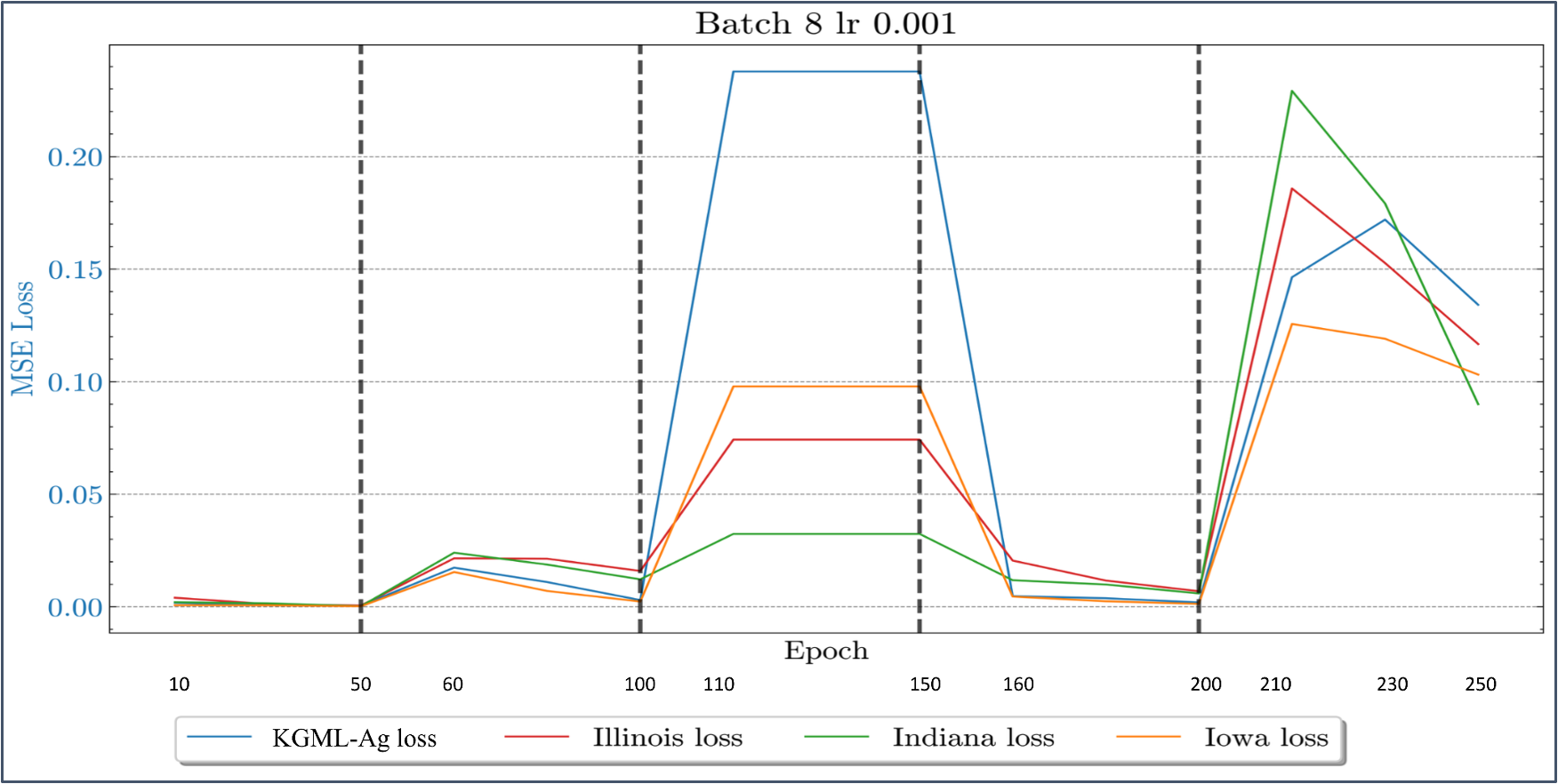}
    \caption{Validation MSE loss evolution across the five-step KGML-ag training process 
    (batch size = 8, learning rate = 0.001). }
    \label{fig:five_step_training}
\end{figure}

To better illustrate the five-stage optimization workflow and model convergence behavior, Figure~\ref{fig:five_step_training} visualizes the validation loss trajectories across the five 
training phases described above. Dashed vertical lines indicate the boundaries of the individual 
phases corresponding to pretraining (Steps 1–2), fine-tuning with real-world data (Step 3 and 5), and finetuning with synthetic data (Steps 4).

\textbf{Spatial Distribution-Shift Aware Knowledge-Guided Machine Learning (SDSA-KGML):} Building on KGML-ag, Sharma et al. (2025) proposed the Spatial Distribution-Shift Aware Knowledge-Guided Machine Learning (SDSA-KGML) framework to explicitly address regional heterogeneity and data distribution shifts. Instead of sharing a single global parameter set $\theta$, SDSA-KGML introduces location-dependent parameters $\theta_{loc}$ that allow each region to learn its own process sensitivities while maintaining shared physics-based constraints\cite{Raissi2019_PINN}.  This design enables the model to capture within-region variability in soil, climate, and management factors, thereby improving local accuracy in heterogeneous agroecosystems across Illinois, Iowa, and Indiana. However, training these region-specific models independently limits their scalability and cross-site generalization, motivating our Level-4 extension (FTBSC-KGML) that augments SDSA-KGML with a global pretraining and local fine-tuning mechanism for efficient site calibration.

\textbf{Spatial Variability Awareness:} To harmonize terminology and position our contribution, we categorize knowledge-guided models by \emph{spatial variability awareness} (Table~\ref{tab:example_table}): 
Level-1 uses location-independent (LI) inputs, outputs, and parameters,
Level-2 uses location-dependent (LD) inputs/outputs with shared global parameters, Level-3 further adopts location-dependent parameters $\theta_{\text{loc}}$, and Level-4 includes transfer-learning. Under this taxonomy, KGML-ag falls into Level-2, SDSA-KGML falls into Level-3, and our \textbf{FTBSC-KGML} is in Level-4.

\begin{table*}[t]
\centering
\small
\caption{Spatial variability awareness levels of knowledge-guided models (LI = Location independent, LD = Location dependent).}
\label{tab:example_table}
\renewcommand{\arraystretch}{1.05}
\setlength{\tabcolsep}{3pt} 
\begin{tabular}{p{0.6cm}p{2.1cm}p{2.6cm}p{1.2cm}p{1.2cm}p{2.0cm}p{2.3cm}p{3.3cm}}
\toprule
\textbf{Level} & \textbf{Taxonomy} & \textbf{Example} & \textbf{Inputs ($x$)} & \textbf{Outputs ($y$)} & \textbf{Parameters ($\theta$)} & \textbf{Transfer learning included} & \textbf{Model representation} \\
\midrule
1 & One size fits all & Global circulation models \cite{stute2001gcm} 

& LI & LI & LI & No 
& \mbox{$y = f_{\text{Knowledge}}\!\left(x;\, \theta\right)$} \\

2 & Spatial explicit & KGML-ag \cite{Liu2024_KGML_Carbon} 
& LD & LD & LI & No 
& \mbox{$y_{\text{loc}} = f_{\text{Knowledge}}\!\left(x_{\text{loc}};\, \theta\right)$} \\

3 & Location specific & SDSA-KGML \cite{Sharma2025SDSA} 
& LD & LD & LD & No 
& \mbox{$y_{\text{loc}} = f_{\text{Knowledge}}\!\left(x_{\text{loc}};\, \theta_{\text{loc}}\right)$} \\

4 & Location calibrated & \textbf{FTBSC-KGML(This work)} 
& LD & LD & LD & Yes 
& \mbox{$y_{\text{loc}} = f_{\text{Knowledge}}\!\left(x_{\text{loc}};\, \theta_{\text{global}\rightarrow\text{loc}}\right)$} \\
\bottomrule
\end{tabular}
\end{table*}

\section{Problem Definition}

The problem is defined as follows:

\noindent\textbf{Input:} \begin{itemize}[leftmargin=*, topsep=0pt, itemsep=0pt, parsep=0pt]
\item Multi-source spatiotemporal covariates (e.g., remote-sensing GPP, climate, soil, management).
\item Locations $l_i$ (site/state) and region labels.
\item Global pretrained weights $\theta^{(0)}$ and physics constraints.
\end{itemize}

\noindent\textbf{Output:} \begin{itemize}[leftmargin=*, topsep=0pt, itemsep=0pt, parsep=0pt]
\item Land-emission value(s) at target site/time (NEE).
\end{itemize}

\noindent\textbf{Objective:} \begin{itemize}[leftmargin=*, topsep=0pt, itemsep=0pt, parsep=0pt]
\item Maximize prediction quality (e.g., minimize MSE).
\end{itemize}

\noindent\textbf{Constraints:} \begin{itemize}[leftmargin=*, topsep=0pt, itemsep=0pt, parsep=0pt]
\item Spatial variability across sites/regions.
\item Data sparsity and uneven spatial coverage.
\item Transfer design: global pretraining $\rightarrow$ local fine-tuning (site calibration).
\item Compute/resource limits.
\end{itemize}

\subsection{Problem Formulation}
The model uses inputs such as Gross Primary Productivity (GPP), climatic variables (e.g., temperature), and soil characteristics. The output is the predicted land emissions, representing the amount of carbon dioxide released by vegetation and soil microorganisms. The main objective is to ensure high predictive accuracy despite challenges posed by spatial variability and limited data availability.

Spatial variability is a major challenge in land-emissions modeling, affecting both inputs and physical processes. Emissions differ sharply across land uses: mountainous regions often have low emissions (sparse vegetation and population), valleys have high emissions (intensive agriculture), and plains have moderate emissions (mixed agricultural and urban activity). This heterogeneity hinders model generalizability and increases the risk of overfitting.

\section{Design Decisions}
This study highlights three key design decisions that differentiate our \textbf{FTBSC-KGML} framework from prior knowledge-guided models. Each decision represents a conceptual choice that shapes the model’s transferability, interpretability, and alignment with physical knowledge.

\begin{itemize}[leftmargin=*, topsep=0pt, itemsep=0pt, parsep=0pt]
    \item \textit{\textbf{Global Pretraining vs. Site-Only Training}}: To assess whether globally shared representations can improve fine-tuning efficiency and local predictive accuracy.
    
    \item \textit{\textbf{With vs. without site-specific calibration}}: Assess whether localized adaptation layers are needed to capture site-dependent process sensitivities.
    
    \item \textit{\textbf{Robustness}}: To examine whether the global-to-local fine-tuning scheme maintains stable performance across different hyperparameter settings (e.g., learning rate, batch size), demonstrating the inherent robustness.
\end{itemize}

These three design axes form the conceptual backbone of our study. Their implementation details are described in the approach section, while their impacts are examined through the sensitivity analysis section.

\section{Proposed Approach}
\label{sec:approach}

We present \textbf{FTBSC-KGML} (\emph{Fine-Tuning-Based Site Calibration for Knowledge-Guided Machine Learning}), 
a framework that extends KGML-ag by introducing cross-site transfer and site-specific calibration under feature heterogeneity.
The proposed method is built upon two complementary ideas:
\begin{enumerate}
    \item \textbf{Global pretraining with site-level fine-tuning:} For knowledge-consistent initialization \& efficient adaptation.
    \item \textbf{Site-specific calibration under feature shifts:} To capture regional variation in climate, soil, and management.
\end{enumerate}
Both ideas are compatible with KGML-ag’s physics-guided constraints (e.g., mass balance, response consistency).

\paragraph{Core Idea I: Global Pretraining then Site-level Fine-tuning.}
Let $D_{\text{global}}=\bigcup_{s\in S}D_s$ denote the aggregated multi-state dataset. 
We first train a knowledge-guided backbone $\theta$ on $D_{\text{global}}$:
\begin{equation}
\theta^{\*}=\arg\min_{\theta}\; L_{\text{pred}}\!\left(D_{\text{global}};\theta\right)+\lambda\,L_{\text{phys}}(\theta),
\end{equation}
where $L_{\text{phys}}$ encodes physics constraints (e.g., mass balance and response consistency).
For a target site $s$, the model is initialized from $\theta^{\*}$ and fine-tuned for local calibration:
\begin{equation}
\theta^{\*}_s=\arg\min_{\theta_s}\; L_{\text{pred}}\!\left(D_s;\theta_s\right)+\lambda\,L_{\text{phys}}(\theta_s)+\mu\,\lVert\theta_s-\theta^{\*}\rVert_2^2.
\end{equation}
This transfer learning procedure transfers cross-site representations of crop–climate–soil interactions while adapting to local heterogeneity,
improving both generalization and convergence in low-sample settings.

\paragraph{Core Idea II: Site-Specific Calibration under Feature Shifts.}
While global pretraining provides a stable, physics-consistent initialization, 
feature distributions differ markedly across regions due to variations in climate, soil, and management practices.
To address these feature shifts, we perform \emph{site-specific calibration} during fine-tuning by introducing lightweight local parameters.
After obtaining the globally informed backbone $f(\cdot;\theta^{\*})$, we attach a calibration head $h_s(\cdot;\phi_s)$ 
(or optional adapter blocks $A_s$) for each site and adapt them using local data $D_s$:
\begin{equation}
\min_{\theta_s,\,\phi_s}\;
L_{\text{pred}}\!\big(D_s;\theta_s,\phi_s\big)
+\lambda\,L_{\text{phys}}\!\big(\theta_s,\phi_s\big)
+\mu\,\lVert\theta_s-\theta^{\*}\rVert_2^2
+\rho\,\lVert\phi_s\rVert_2^2.
\end{equation}
Here $\phi_s$ denotes the site-specific calibration parameters, 
$\theta_s$ are lightly fine-tuned global weights initialized from $\theta^{\*}$, 
and $\rho$ regularizes the capacity of local adaptation to prevent overfitting.
This calibration mechanism preserves the generalizable structure learned during pretraining
while allowing flexible adaptation to local feature distributions.

 As summarized in Equations~(1)–(3), the framework consists of two consecutive phases: 
a \textit{global pretraining phase} that learns physics-consistent, transferable representations from aggregated multi-state data, 
and a \textit{site-calibrated fine-tuning phase} that adapts these representations to the local feature distributions of each site or state. 
This hierarchical design allows the model to retain global generalization capability while improving local accuracy under spatial heterogeneity.

\begin{figure}[t]
    \centering
    \includegraphics[width=0.92\linewidth]{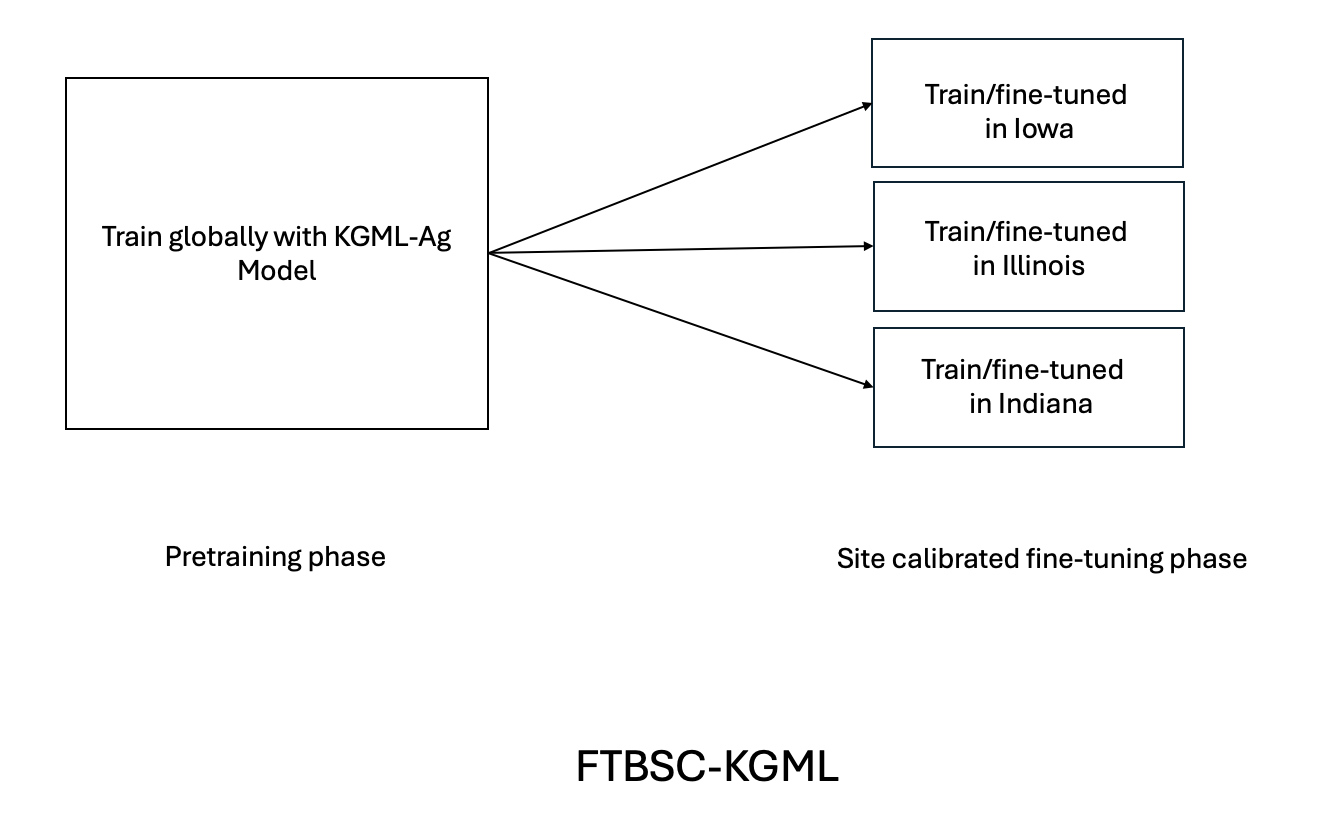}
    \caption{Overview of the proposed \textbf{FTBSC-KGML} framework.}
    \label{fig:intro}
\end{figure}

Figure~\ref{fig:intro} conceptually illustrates how FTBSC-KGML operationalizes the ``global-to-local'' transfer-learning paradigm. 
The global pretraining stage provides a robust, knowledge-guided initialization that captures cross-site interactions and ensures physical realism. 
while the site-specific fine-tuning stage introduces localized adjustments that account for spatial variability in climate, soil, and management conditions. 
Together, these stages establish a balanced framework that achieves both generalization and specialization, outperforming purely global or purely site-specific approaches.

\paragraph{Complementarity of the two ideas} Idea~I supplies a globally informed, physically consistent prior that stabilizes optimization across sites \cite{Karpatne2017_TGDS,Karniadakis2021_PIML}. Idea~II explicitly models cross-site feature shifts through site-specific calibration, enhancing locality without sacrificing transferability. Together, they balance global generalization and local specialization: global pretraining prevents overfitting under sparse data,
while per-site calibration improves fidelity to regional agroecosystem patterns. Empirically, FTBSC-KGML achieves lower validation error and faster convergence than both state-only (\emph{Non-Transfer Learning}) and global-only (\emph{no specialization}) baselines.

\section{Experimental Evaluation}

\textbf{Experimental Goals:} Our evaluation is organized around two categories of research questions:
(1) \textbf{Comparative Analysis} and (2) \textbf{Sensitivity Analysis}. 
These questions define the scope of the experiments and guide the analysis of spatial heterogeneity and global-to-local transfer learning.

\textbf{Comparative Analysis:} (1) How do local models (site-only, SDSA-KGML) compare with the global KGML-ag model in predictive accuracy? (2) How well do models trained in one region transfer to another? (3) How does site-level calibration affect a globally pretrained model?

\textbf{Sensitivity Analysis:} (1) How sensitive is the proposed model to changes in hyperparameters such as learning rate and batch size? (2) How stable is the model across states with different levels of data sparsity and feature heterogeneity?

\subsection{Data and Implementation}
We built on the multi-source dataset introduced by the KGML-ag framework~\cite{Liu2024_KGML_Carbon}, 
which integrates climate, soil, productivity, and management information across Midwestern U.S. croplands. 
While the original KGML-ag dataset provides harmonized inputs for knowledge-guided modeling, 
we reorganized and partitioned these data into \textbf{site-level subsets} corresponding to Illinois, Iowa, and Indiana 
to enable the evaluation of our \textbf{ FTBSC-KGML} framework. The data quantity ratio in Iowa, Indiana, and Illinois is 41:27:22. The data sources included:
\begin{itemize}[leftmargin=*, topsep=0pt, itemsep=0pt, parsep=0pt]
    \item \textbf{Climate:} NLDAS-2 daily temp., precip., and rad;
    \item \textbf{Soil:} gSSURGO gridded soil properties;
    \item \textbf{Productivity:} gross primary productivity (GPP);
    \item \textbf{Management:} yearly crop yield: NASS;
    \item \textbf{Synthetic biophysical data:} 18-year \textit{ecosys} simulations used for knowledge-guided pretraining .
\end{itemize}

All preprocessing procedures (normalization, spatial alignment, and temporal aggregation) 
followed the KGML-ag data pipeline~\cite{Liu2024_KGML_Carbon}. 
Our contribution lies in constructing \textbf{state-specific subsets} to support 
cross-site pretraining and fine-tuning experiments, 
allowing direct evaluation of knowledge transfer and spatial calibration across heterogeneous agroecosystem regions.

\subsection{Evaluation Metrics}

Model performance was evaluated using \textbf{mean squared error (MSE)} on validation sets, comparing predicted versus observed fluxes and yields. 
A smaller MSE indicates better predictive accuracy, as it measures the average squared deviation between predictions and observations. 
Thus, a reduction in MSE reflects improved accuracy and robustness under spatial variability. 
We also computed the \textbf{relative MSE improvement} of FTBSC-KGML over site-only baselines, SDSA-KGML \cite{Sharma2025SDSA}, to quantify the benefit of transfer learning.

\paragraph{Design matrix of training regimes.}
We organize the comparison in a $2\times2$ matrix, distinguishing \emph{cross-site} from \emph{site-specific} information (Table~\ref{tab:site_cross_taxonomy}). 
The YES/YES quadrant corresponds to our \textbf{FTBSC-KGML}---global knowledge-guided pretraining followed by site-level fine-tuning. 
The other quadrants cover site-only models (such as SVANN \cite{gupta2021spatial}), a site-independent one-size-fits-all baseline, and an empty reference.

\begin{table}[htbp]
\centering
\scriptsize
\caption{Taxonomy of learning strategies leveraging site-specific and cross-site information.}
\label{tab:site_cross_taxonomy}
\resizebox{\linewidth}{!}{
\begin{tabular}{|p{2.8cm}|c|c|}
\hline
 & \multicolumn{2}{c|}{\textbf{Cross-site}} \\ \cline{2-3}
 & \textbf{No} & \textbf{Yes} \\ \hline
\textbf{Site-specific} & & \\ \hline
No & None & \textit{Global (OSFA)} \\ \hline
Yes & \textit{SVANN} & \textbf{FTBSC-KGML} \\ \hline
\end{tabular}
}
\end{table}

\subsection{Comparative Analysis}

\subsubsection{Experimental Design}

We compare four complementary training configurations:

\begin{itemize}
    \item \textbf{Site-only training}: Independent models trained from scratch for each state (Illinois, Iowa, Indiana) without cross-state knowledge transfer.

    \item \textbf{Global-only training}: A single model trained on the aggregated multi-state dataset and applied directly to each state without local adaptation.

    \item \textbf{Cross-regional prediction}: Models trained in one state are evaluated on other states to assess transferability under distribution shift.

    \item \textbf{Global$\rightarrow$Site Fine-tuning (FTBSC-KGML, proposed)}: A global model pretrained on all states is fine-tuned separately for each state to achieve localized calibration under spatial heterogeneity.
\end{itemize}

This comparison isolates the impact of global pretraining and site-level calibration by keeping the architecture and physics-guided constraints constant. The experiments evaluate both daily carbon flux variables (Ra, Rh, Reco (ecosystem respiration, Ra+Rh), NEE) and annual crop yield, capturing key aspects of carbon-cycle behavior in agroecosystems.

\subsubsection{Experimental Results of Comparative Analysis}

\subsubsection{How do local (site-only) and global models compare in predictive accuracy?}

We first compared \emph{site-only} models, SDSA-KGML \cite{Sharma2025SDSA}, to a \emph{global-only} model.
The site-specific KGML-ag models were trained independently on data from Iowa and Indiana and compared with a single KGML-ag model trained on the combined dataset ( Iowa + Indiana). This setup contrasts localized models with a regional model integrating all states.

Our findings show that KGML-ag models trained on state-specific data often achieve lower MSE on their respective states than the purely global model, highlighting the benefit of location-based parameters in capturing spatial heterogeneity. 
However, these gains in local accuracy come at the cost of reduced cross-site generalization: models trained in one state degrade when applied to others, indicating a bias–variance trade-off between local fidelity and spatial transferability.

\begin{figure}[t]
    \centering
    \includegraphics[width=0.7\linewidth,height=0.6\linewidth]{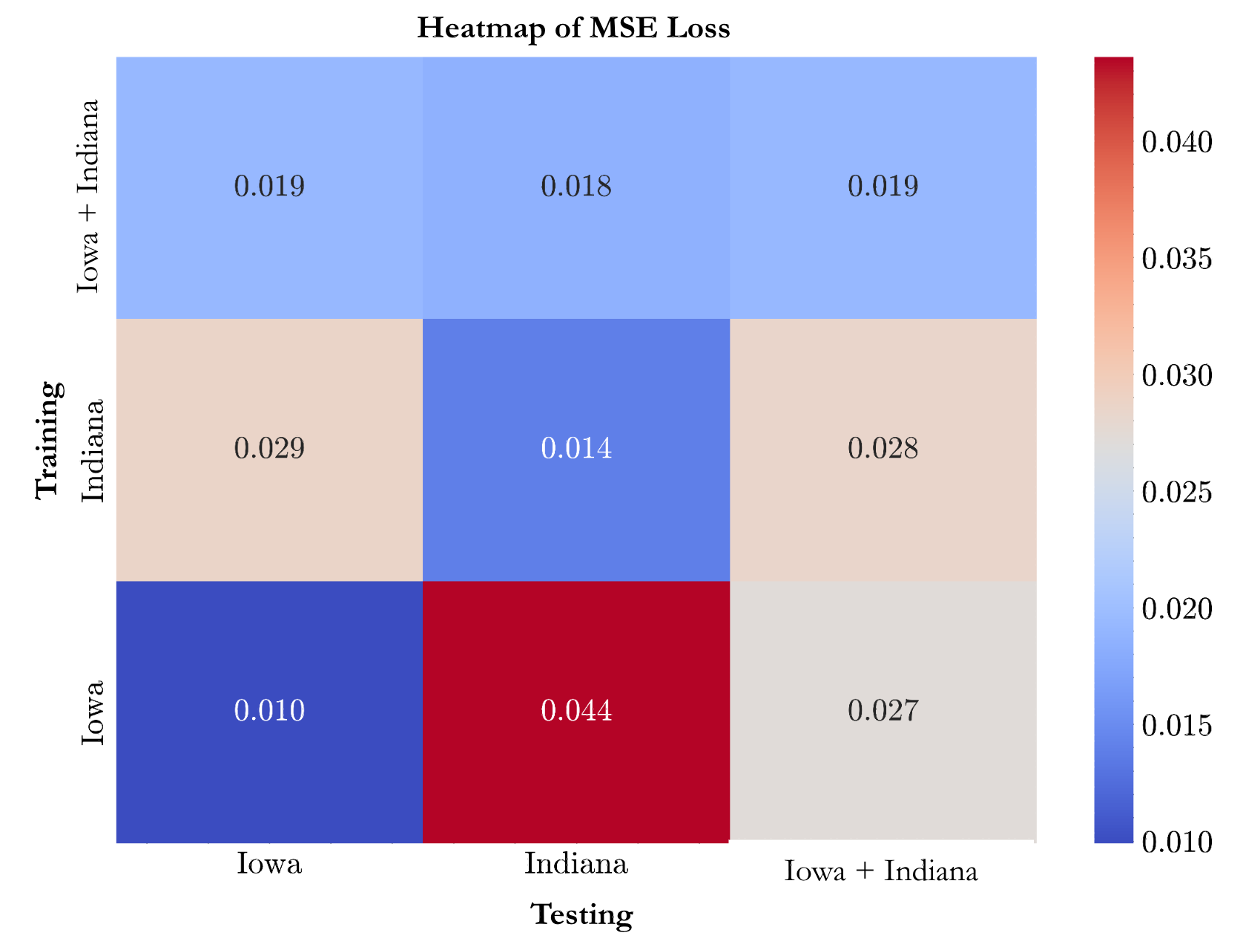}    
    \caption{Heatmap of validation MSE across different state-level training and testing combinations.}
    \label{fig:exp123}
\end{figure}

Figure~\ref{fig:exp123} visualizes a heatmap of MSE across different training–testing combinations. 
Bluer colors correspond to lower MSE values. 
The most significant errors occur when a model trained on Iowa is tested on Indiana, underscoring substantial regional differences. This motivates the addition of a \emph{global-to-local} mechanism that combines the strengths of both global and local strategies.

\textbf{How well do models trained in one region transfer to other regions? :} We train KGML-ag models in Indiana, Iowa, and Indiana+Iowa using state-specific weights and applying them to other states. Due to limited data in Illinois, we only used Iowa and Indiana data for this question. Results confirm that localizing KGML-ag to a specific state and evaluating on the same state yields substantially lower MSE and higher predictive accuracy.  However, when these localized models are applied cross-regionally, performance degrades, indicating limited spatial generalization.  With location-specific parameters, SDSA-KGML captures within-state variability more faithfully and produces estimates tailored to local conditions, but at the expense of robustness to unseen regions. Conversely, KGML-ag with location-independent parameters achieves stronger cross-region transfer but at the expense of state-specific accuracy. These results underscore the need for a framework like FTBSC-KGML that balances local calibration with transferable global knowledge.

\textbf{How does site-level calibration affect a globally pretrained model? :} We then evaluated \textbf{FTBSC-KGML}, which combines global pretraining with site-level fine-tuning.  A global KGML-ag model was first pretrained on the combined dataset from Illinois, Iowa, and Indiana; the resulting weights were then fine-tuned separately for each state. This setting assesses generalization from aggregated training to state-specific testing and characterizes the model’s adaptability under different agricultural and meteorological conditions.

\begin{figure}[htbp]
    \centering
    \includegraphics[width=0.85\linewidth]{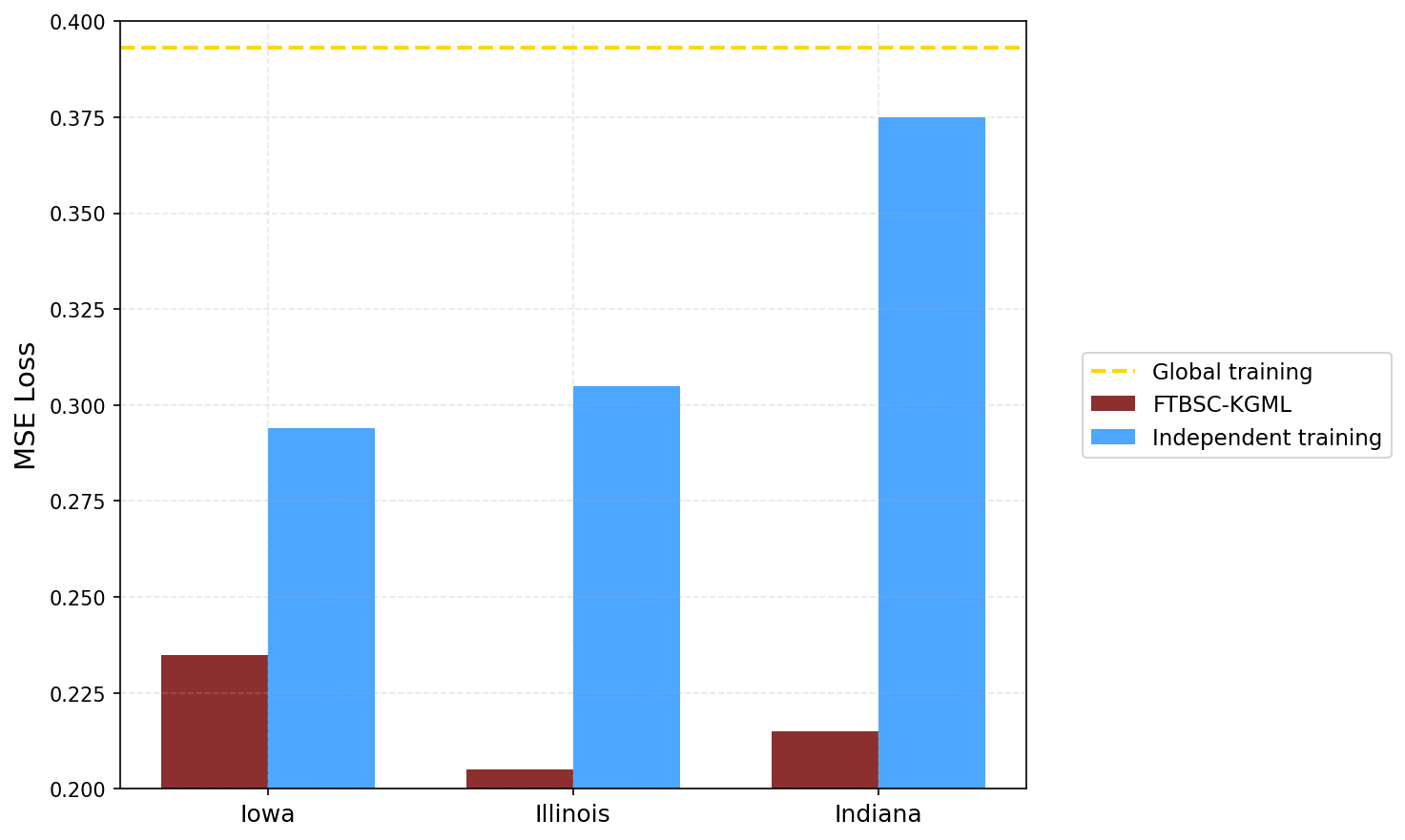}
    \caption{Validation MSE by state comparing \textbf{with global pretraining} (global $\rightarrow$ state fine-tuning; maroon) vs.\ \textbf{without pretraining} (state-only training from scratch; blue). Pretraining yields consistently lower MSE in IA/IL/IN.}
    \label{fig:pretrain_vs_stateonly}
\end{figure}

Figure~\ref{fig:pretrain_vs_stateonly} compares validation MSE for \emph{global pretraining + fine-tuning} versus \emph{state-only training from scratch}. We observe consistently lower validation MSE with pretraining across Iowa, Illinois, and Indiana (\emph{Iowa: $\Delta$MSE $= 0.050$ (17.0\%); Illinois: $\Delta$MSE $= 0.095$ (31.2\%); Indiana: $\Delta$MSE $= 0.163$ (43.6\%)}). Global pretraining learns transferable crop–climate–soil representations and reduces the optimization burden during state-level fine-tuning, thereby mitigating overfitting when state data are limited. 

These results support our claim that \textbf{site calibration atop shared pretraining} improves local accuracy while preserving knowledge-guided behavior. FTBSC-KGML outperforms global-only, site-only, and cross-regional strategies, especially in data-limited states such as Illinois and Indiana, where it remains stable and achieves larger relative gains.

\textbf{Comparative Analysis Discussion:} Across all evaluations, the KGML-based emulator exhibits strong fidelity to the process-based model while maintaining physical consistency.  Global pretraining establishes transferable ecological structure, and fine-tuning provides localized correction without introducing instability. The performance gains are most significant in data-limited states, showing that shared representations are particularly valuable when local observations are scarce. The comparisons among site-only, global-only, cross-regional, and FTBSC-KGML regimes reveal that neither purely local nor purely global strategies are sufficient in heterogeneous agroecosystems.  FTBSC-KGML effectively interpolates between these extremes by leveraging global pretraining to provide a strong initialization and a physics-informed structure, followed by site-specific calibration to capture local patterns. The physics consistency analysis further supports that this adaptation remains knowledge-consistent, which is critical for downstream scientific use.

\textbf{Sensitivity Analysis:} \label{sec:sensitivity} In this phase, we investigated the robustness of the proposed framework under variations in two key hyperparameters—\textit{learning rate (lr)} and \textit{batch size (bs)}. Hyperparameter sensitivity is critical for knowledge-guided models that combine physics constraints and data-driven learning, as training instabilities can distort the delicate balance between physical consistency and empirical adaptation. We examine whether the transfer-learning advantages remain valid when these hyperparameters are perturbed, and whether any states exhibit distinct responses due to data heterogeneity.

\begin{figure*}[ht!]
    \centering
    \includegraphics[width=0.32\linewidth]{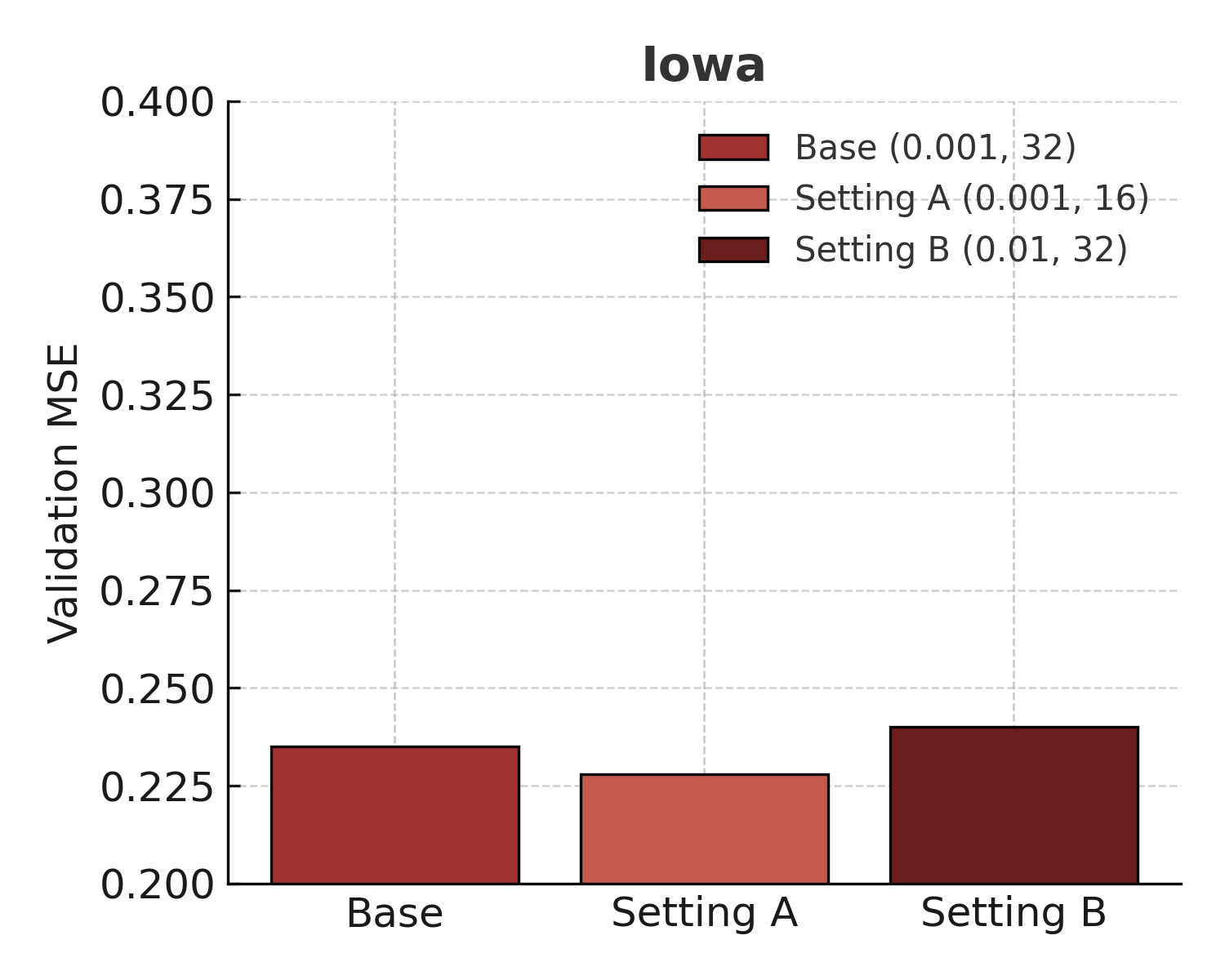}
    \includegraphics[width=0.32\linewidth]{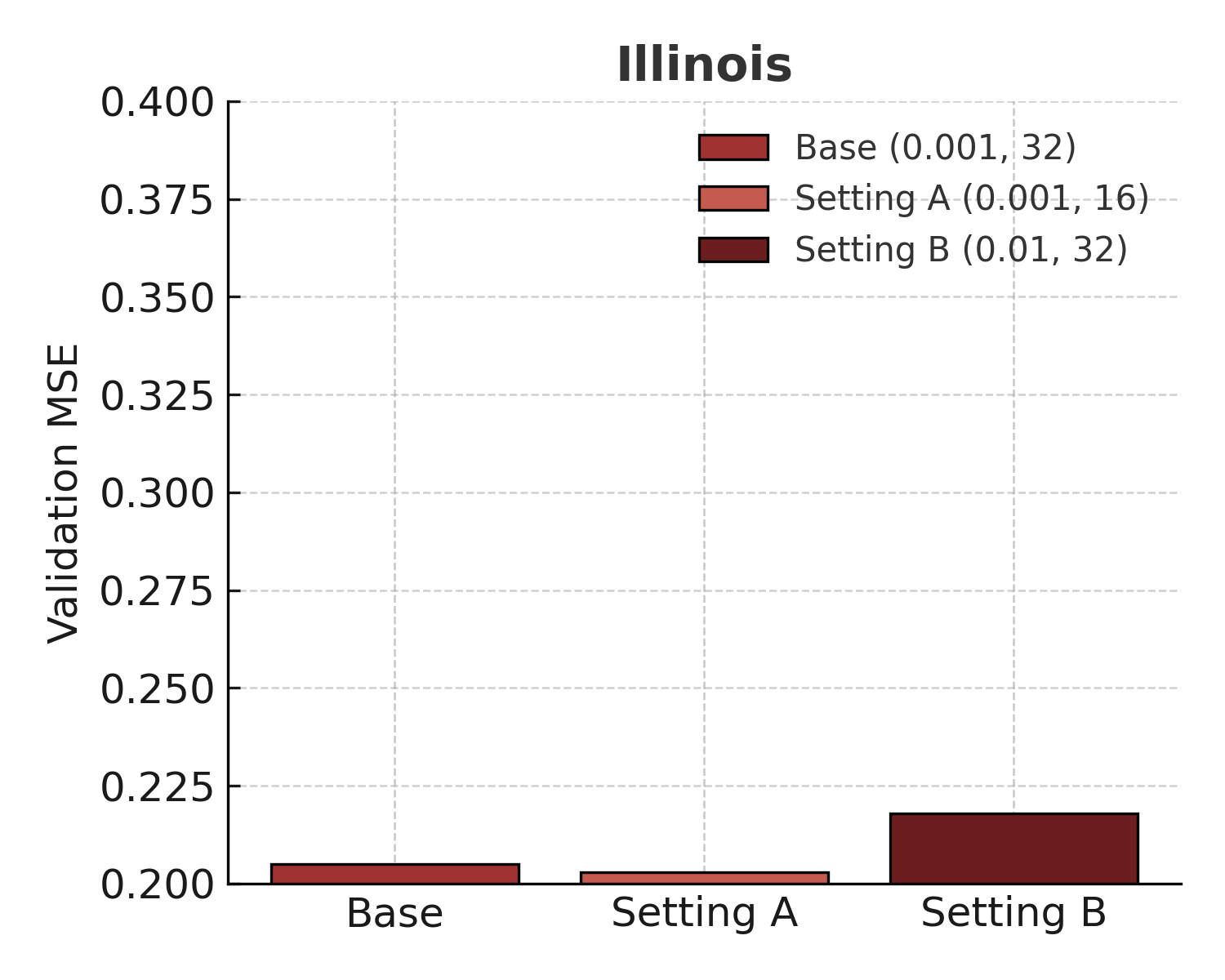}
    \includegraphics[width=0.32\linewidth]{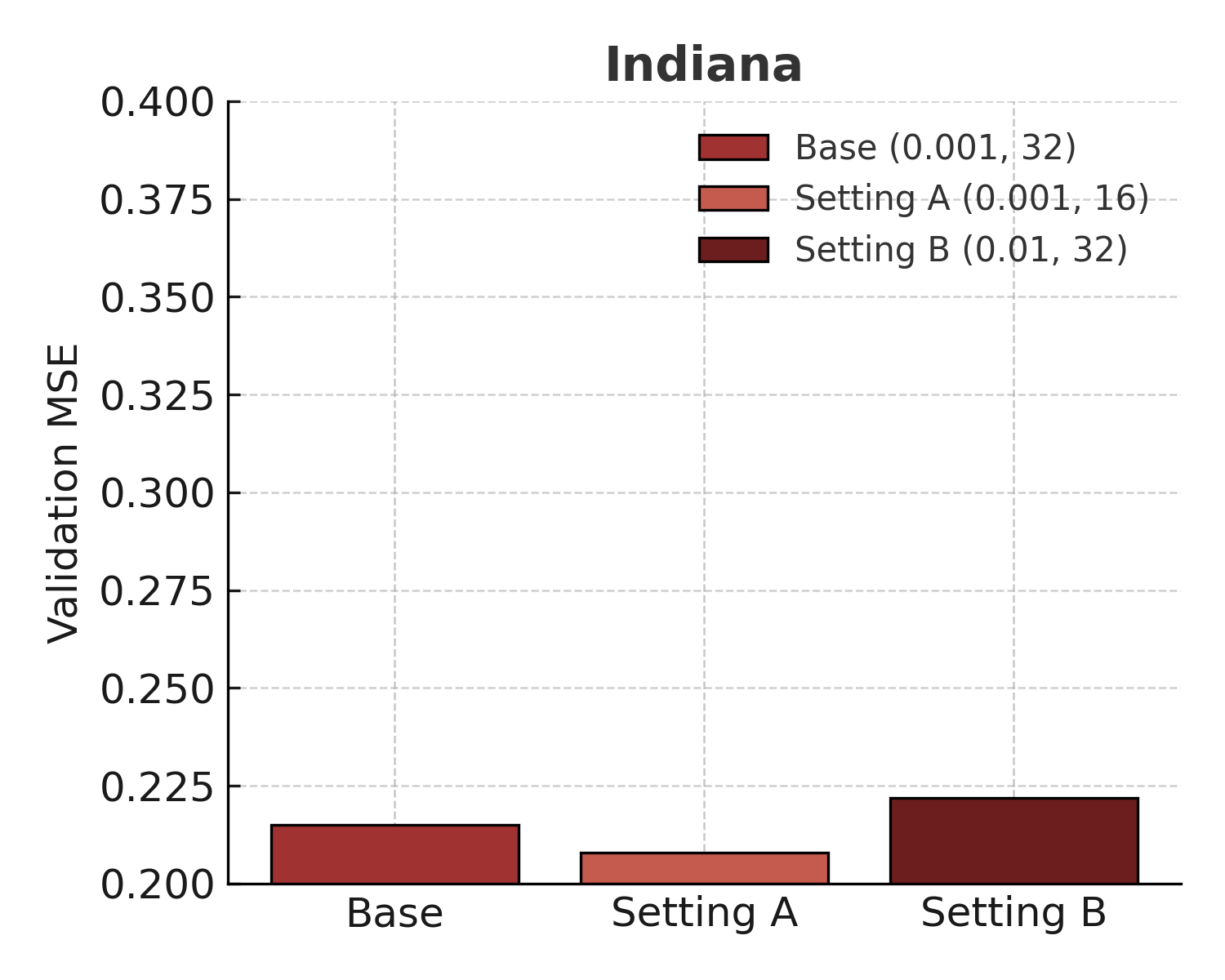}
    \caption{\textbf{Validation MSE across three hyperparameter settings for (a) Iowa, (b) Illinois, and (c) Indiana.} 
    Bars show results for the baseline ($\text{lr}=0.001, \text{bs}=32$), reduced batch size ($\text{lr}=0.001, \text{bs}=16$), and increased learning rate ($\text{lr}=0.01, \text{bs}=32$). 
    Lower values indicate better predictive accuracy. 
    Across all states, the proposed T-learning models remain stable and consistently achieve low MSE under varying training conditions.}
    \label{fig:sens_IA_IL_IN}
\end{figure*}

\textbf{Experimental Design: } The original experiments used a learning rate of 0.001 and a batch size of 32. To evaluate sensitivity, we introduced two perturbed configurations:
\begin{itemize}
    \item \textbf{Setting A:} $\text{lr}=0.001$, $\text{bs}=16$ (smaller batch size);
    \item \textbf{Setting B:} $\text{lr}=0.01$, $\text{bs}=32$ (larger learning rate).
\end{itemize}
Each configuration followed the same 5-step training scheme as the main experiment, including pretraining on aggregated multi-state data followed by site-level fine-tuning on Illinois, Iowa, and Indiana datasets. 

\subsubsection{How does the model respond to reducing the batch size during training?}
We first trained the FTBSC-KGML models with $\text{lr}=0.001$ and $\text{bs}=16$, halving the batch size from the original setup.
Smaller batch sizes generally increase gradient variance and can expose instabilities in models that depend on physics-based regularization.
However, as shown in Figure~\ref{fig:sens_IA_IL_IN}(a)--(c), the validation MSE slightly \textit{decreased} in all three states.
Specifically, Iowa achieved a 6.8\% lower validation loss, Illinois 4.3\%, and Indiana 3.1\%.

The smaller batches may have enhanced diversity within each gradient step, allowing the model to better capture local feature distributions without overfitting to global patterns.

Interestingly, the relative difference between states remains consistent with previous results: Iowa achieves the lowest overall error, while Illinois and Indiana exhibit similar trajectories.
This mirrors the pattern observed in comparative analysis, where data volume (Iowa: 41\% of total) played a dominant role in reducing variance.
Hence, the reduction in MSE under smaller batches reflects improved convergence rather than overfitting, and confirms that the FTBSC-KGML training pipeline remains stable under higher stochastic noise.

\subsubsection{How does the model respond to increasing the learning rate?}
We next examined the effect of increasing the learning rate by an order of magnitude to $\text{lr}=0.01$ while keeping $\text{bs}=32$.
This test probed the upper bound of optimization stability.
Validation MSE values remained low across all three states and stayed below the original independent training baselines reported in Figure~\ref{fig:pretrain_vs_stateonly}.
This suggests that the global pretraining stage provides a stable initialization that resists divergence even under more aggressive learning updates.

\subsubsection{How stable is the model across states with different levels of data sparsity and feature heterogeneity?}
Across both experiments, the absolute variation in MSE due to hyperparameter perturbations was smaller than the variation observed across states in the comparative analysis.
This indicates that the model’s sensitivity to data heterogeneity—differences in climate, soil, and management—is substantially higher than its sensitivity to training configuration.
Such robustness is a desirable property for large-scale geospatial modeling, where computational limits or site-specific data availability often require nonuniform batch sizes or adaptive learning rates.

\textbf{Sensitivity Analysis Discussion:} The resilience of FTBSC-KGML to learning-rate and batch-size changes can be attributed to its two-stage training architecture. The global pretraining step serves as meta-regularization, producing well-conditioned representations that encode physical knowledge (e.g., carbon balance, soil–climate interactions). Fine-tuning then performs small, localized parameter updates that adapt these representations to site-level distributions. Thus, while standard neural networks often exhibit high sensitivity to learning rate and batch size, the proposed framework maintains a balance between transferability and locality, effectively “anchoring” the optimization trajectory to a physically meaningful manifold. Moreover, the presence of the $L_{phys}$ term in the loss function constrains deviations during fine-tuning, preventing the model from drifting toward overfitted minima. In this way, hyperparameter perturbations mainly affect the rate of convergence rather than the final performance, which explains the consistent validation losses across settings. This behavior mirrors that of well-regularized meta-learning models, in which learned initializations dominate the optimization dynamics.

Overall, the sensitivity analysis shows that the benefits of global pretraining and site-level calibration are insensitive primarily to hyperparameters. Moderate changes in learning rate and batch size produce only small shifts in validation MSE, confirming the robustness and reproducibility of the transfer-learning setup. The model maintains low errors across all states, even under less favorable optimization settings, highlighting two key implications:
\begin{enumerate}
    \item \textbf{Generalization robustness:} The transfer-learning design yields consistent gains across hyperparameters, enabling reliable deployment under varied computational settings.
    \item \textbf{Stability through knowledge guidance:} Embedded physics constraints provide an inductive bias that regularizes gradient updates and prevents divergence.
\end{enumerate}
\section*{Conclusion and Future Work}

This work extends the SDSA-KGML framework with Fine-Tuning-Based Site Calibration (FTBSC-KGML), which adapts globally pretrained knowledge-guided models to local sites. Across multiple states, FTBSC-KGML consistently achieves lower validation error than site-only training, showing that global pretraining followed by local fine-tuning strikes a better balance between cross-site generalization and site-specific accuracy. Fine-tuning global models with site-level data yields better adaptation to local spatial variability than training each site independently, especially on data-limited sites, where the method remains stable and outperforms site-only baselines.

\textbf{Future Work:} Building on the SDSA-KGML design, this study reaffirms that location-dependent parameters improve the model’s ability to capture spatial heterogeneity\cite{ghosh2024towardskriging}. The proposed FTBSC-KGML framework bridges global learning and local calibration, offering a practical route to scalable, spatially aware modeling. However, physics-based constraints  \cite{yang2025geo,sharma2020analyzing,sharma2022abnormal,sharma2022towards,sharma2024physics,sharma2022spatiotemporal,kumar2015graph} are intended to regularize fine-tuning; a full quantitative assessment of whether site-level adaptation preserves the physical consistency learned during global pretraining remains open. Future work will compute and compare the physics-consistency loss $L_{\text{phys}}$ to measure mass-balance and non-negativity violations for both the global model and the site-specific fine-tuned models, to detect any drift in physical realism, and to motivate explicitly physics-preserving calibration strategies \cite{li2023eco,sharma2025towards,sharma2025towards1,yang2024towards,sharma2022understanding,sharma2024physics}. Future methodological work will focus on: (1) neighborhood-based calibration to replace static region boundaries; and (2) parameter-efficient, domain-adaptive fine-tuning for broader scalability. To enhance robustness, we will explore dynamic learning-rate schedules (e.g., cosine decay, adaptive optimizers) and parameter-efficient schemes (e.g., adapters \cite{houlsby2019adapter}, LoRA \cite{hu2022lora}) to improve transfer efficiency further. Overall, our results show that FTBSC-KGML is both numerically and conceptually robust, maintaining superior performance across sites and optimization settings.



\bibliography{refs}

\end{document}